\documentclass{article}

\usepackage[accepted]{icml2018}

\usepackage[toc,page]{appendix}
\usepackage{color}

\usepackage{amsmath}
\usepackage[utf8]{inputenc} 
\usepackage[T1]{fontenc}    
\usepackage{hyperref}       
\usepackage{url}            
\usepackage{booktabs}       
\usepackage{amsfonts}       
\usepackage{nicefrac}       
\usepackage{microtype}      
\usepackage{amsmath}
\usepackage{amsthm}
\usepackage{amssymb}
\usepackage{graphicx}

\newtheorem{mydef}{Definition}

\DeclareMathOperator*{\argmin}{arg\,min}

\newcommand{\se}[1]{{\textcolor{black}{#1}}}
\newcommand{\secor}[2]{{\textcolor{black}{#2}}}
\newcommand{\secmt}[1]{}

\newcommand{\ie}{\emph{i.e. }}

\definecolor{mypink1}{rgb}{0.858, 0.188, 0.478}

\definecolor{orange}{rgb}{1,0.5,0}
\newcommand{\al}[1]{{\textcolor{black}{#1}}}

\newcommand{\bmL}{\mathcal{L}}
\newcommand{\bmH}{\mathcal{H}}

\newcommand{\X}{\mathcal{X}}

\newcommand{\inspace}{\mathcal{X}}

\newcommand{\Y}{\mathcal{Y}}
\newcommand{\V}{\mathcal{V}}

\newcommand{\Edge}{\mathcal{E}}
\newcommand{\G}{\mathcal{G}}
\newcommand{\Yr}{\mathcal{Y}_r}

\newcommand{\hx}{h(x)}
\newcommand{\rx}{r(x)}
\newcommand{\pred}{f^{h,r}}

\newcommand{\predh}{\hat{h}(x)}
\newcommand{\predr}{\hat{r}(x)}
\newcommand{\lossa}{\Delta_{a}}
\newcommand{\lossabin}{\Delta_{a}^{bin}}
\newcommand{\truerisk}{R}

\newcommand{\mapa}{\psi_a}
\newcommand{\mapwa}{\psi_{wa}}
\newtheorem{theorem}{Theorem}

\begin{document}

\twocolumn[
\icmltitle{Structured Output Learning with Abstention: Application to Accurate Opinion Prediction}




\begin{icmlauthorlist}
\icmlauthor{Alexandre Garcia}{to}
\icmlauthor{Slim Essid}{to}
\icmlauthor{Chlo\'e Clavel}{to}
\icmlauthor{Florence d'Alch\'e-Buc}{to}
\end{icmlauthorlist}

\icmlaffiliation{to}{LTCI, Telecom ParisTech, Paris, France}

\icmlcorrespondingauthor{Alexandre Garcia}{algarcia@enst.fr}

\icmlkeywords{Machine Learning, ICML}

\vskip 0.3in
]

\printAffiliationsAndNotice{}




%
\begin{abstract}
Motivated by Supervised Opinion Analysis, we propose a novel framework devoted to Structured Output Learning with Abstention (SOLA). The structure prediction model is able to abstain from predicting some labels in the structured output at a cost chosen by the user in a flexible way. For that purpose, we decompose the problem into the learning of a pair of predictors, one devoted to structured abstention and the other, to structured output prediction. To compare fully labeled training data with predictions potentially containing abstentions, we define a wide class of asymmetric abstention-aware losses. Learning is achieved by surrogate regression in an appropriate feature space while prediction with abstention is performed by solving a new pre-image problem. Thus, SOLA extends recent ideas about Structured Output Prediction via surrogate problems and calibration theory and enjoys statistical guarantees on the resulting excess risk. Instantiated on a hierarchical abstention-aware loss, SOLA is shown to be relevant for fine-grained opinion mining and gives state-of-the-art results on this task. Moreover, the abstention-aware representations can be used to competitively predict user-review ratings  based on a sentence-level opinion predictor.
\end{abstract}

\section{Introduction}
Up until recent years, opinion analysis in reviews has been commonly handled as a supervised polarity (positive vs. negative) classification problem. However, understanding the grounds on which an opinion is formed 
is of highest interest for decision makers. Aligned with this goal, the emerging field of aspect-based sentiment analysis \cite{pontiki2016semeval} has evolved towards a more involved machine learning task where opinions are considered to be structured objects---typically hierarchical structures linking polarities to aspects and relying on different units of analysis (\ie sentence-level and review-level) as in \cite{marcheggiani2014hierarchical}. While this problem has attracted 
a growing attention from the structured output prediction community, it has also raised an unprecedented challenge: the human interpretation of opinions expressed in the reviews is subjective and the opinion aspects and their related polarities are sometimes expressed in an ambiguous way and difficult to annotate \cite{clavel2016sentiment,marcheggiani2014hierarchical}. 
In this context, the prediction error should be flexible and should integrate this subjectivity so that, for example, mistakes on one aspect do not interfere with the prediction of polarity.

In order to address this issue, we propose a novel framework called Structured Output Learning with Abstention (SOLA) \secor{to make the system to abstain}{which allows for abstaining} from predicting parts of the structure\secor{. This flexibility enables the prediction system}{, so as} to avoid providing erroneous insights about the object to be predicted, therefore increasing \secor{its}{} reliability. The new approach extends the principles of learning with abstention recently introduced for binary classification \cite{CortesAbstention} and generalizes surrogate least-square loss approaches to Structured Output Prediction recently studied in \cite{brouard2016input,Ciliberto,osinski2017}. The main novelty comes from the \secor{definition}{introduction} of an asymmetric loss, based on embeddings of desired outputs and \secor{predicted outputs}{outputs predicted} with abstention in the same space.  Interestingly, similarly to the case of Output Kernel Regression \cite{brouard2016input} and appropriate inner product-based losses \cite{Ciliberto}, the approach relies on a simple surrogate \se{formulation, namely a} least-squares \secor{problem}{formulation} followed by the resolution of a \secor{novel}{new} pre-image problem.
The paper is organized as follows. Section 2 introduces the problem to solve and the novel framework, SOLA. Section 3 provides statistical guarantees about the excess risk in the framework of Least Squares Surrogate Loss while section 4 is devoted to the pre-image developed for hierarchical output structures. Section 5 presents the numerical experiments and Section 6 draws a conclusion.

\section{Structured Output Labeling with Abstention}
\label{SOLA}
Let $\X$ be the input sample space. We assume a target graph structure of interest, $\G = (\V=\{\nu_1, \ldots, \nu_d\}, \Edge: \V \times \V \to \{0,1\})$ where $\V$ is the set of vertices and $\Edge$ is the edge relationship between vertices. A legal {\it labeling} or {\it assignment} of $\G$ is a $d$-dimensional binary vector, $y \in \{0,1\}^d$, that also satisfies some properties induced by the graph structure, \ie by $\Edge$. We call $\Y$ the subset of $\{0,1\}^d$ that contains all possible legal labelings of $\G$. Given $\G$, the goal of Structured Output Labeling is to learn a function $f: \X \to \Y$ that predicts a legal labeling $\hat{y}$ given some input $x$. Let us emphasize that $x$ does not necessarily share the same structure $\G$ with the outputs objects. For instance, in Supervised Opinion Analysis, the inputs are reviews in natural language described by a sequence of feature vectors, each of them representing a sentence. 
Extending Supervised Classification with Abstention \cite{CortesAbstention}, Structured Output Learning with Abstention aims at learning a pair of functions $(h,r)$ from $\X$ to $Y^{H,R} \subset \{0,1\}^d \times \{0,1\}^d$ composed of a predictor $h$ that predicts the label of each component of the structure and an abstention function $r$ that determines on which components of the structure $\G$ to abstain from predicting a label. If we note $\Y^\star \subset \{0,1,a \}^d$, the set of legal labelings with abstention where $a$ denotes the abstention label, then the abstention-aware predictive model $\pred: \inspace \rightarrow \Y^\star$ is defined from $h$ and $r$ as follows:
\begin{align}\label{predictor}
\pred(x)^T &= [ \pred_1(x),\ldots, \pred_d(x)], \nonumber\\ \pred_i(x) &= 1_{\hx_i=1}1_{\rx_i=1} + a 1_{\rx_i=0}.
\end{align}
Now, assuming we have 
a random variable $(X,Y)$ taking its values in $\X \times \Y$ and distributed according to a probability distribution $\mathcal{D}$. Learning the predictive model raises the issue of designing an appropriate abstention-aware loss function to define a learning problem as a risk minimization task. Given the relationship in Eq. \secor{\ref{predictor}}{\eqref{predictor}}, a risk on $\pred$ can be converted into a risk on the pair $(h,r)$ using an abstention-aware loss $\lossa : \Y^{H,R} \times \Y \to \mathbb{R}^+$:
\begin{equation}\label{truerisk}
\truerisk(h,r) = \mathbb{E}_{x,y \sim \mathcal{D}} \; \lossa( \hx,\rx,y ).
\end{equation}
In this paper, we propose a family of abstention-aware losses that both generalizes the abstention-aware loss in the binary classification case (see \cite{CortesAbstention}) and extends the scope of hierarchical losses previously proposed by \cite{cesa2006hierarchical} for Hierarchical Output Labeling tasks. 
An abstention-aware loss is required to deal asymmetrically with observed labels which are supposed to be complete and predicted labels which may be incomplete due to partial abstention. We thus propose the following general form for the $\lossa$ function:
\begin{align}\label{lossadef}
\lossa(\hx,\rx,y) =\langle \mapwa(y), C \mapa(\hx,\rx) \rangle,
\end{align}
 

relying on a bounded linear operator (a rectangular matrix) $C: \mathbb{R}^p \to \mathbb{R}^q$ and two bounded feature maps: $\mapa: \Y^{H,R} \to \mathbb{R}^p$ 
devoted to outputs with abstention and $\psi_{wa}: \Y  \to \mathbb{R}^q$, devoted to outputs without abstention. The three ingredients of the loss $\lossa$ must enable the loss to be \secor{positive}{non negative}. This is the case for the following examples.

In {\bf Binary classification with abstention}, we have $\Y = \{0,1\}$ and the abstention-aware loss $\lossabin$ is defined by :
\begin{equation*}
\lossabin(\hx,r(x),y) = \begin{cases}
               1 \text{ if } y\neq \hx \text{ and } r(x) = 1\\
               0 \text{ if } y = \hx \text{ and } r(x) = 1\\
               c \text{ if } r(x) = 0
            \end{cases},
\end{equation*}
where $c \in [0,0.5]$ is the rejection cost; with $\rx = 0$, in case of abstention and $1$, otherwise.
This can be written with the corresponding functions $\mapwa$ and $\mapa$ defined as: 
\begin{align*}
\mapwa(y) &= \begin{pmatrix} 
y \\
1-y
\end{pmatrix}, \;
C = \begin{pmatrix} 
0 & 1 & c  \\
1 & 0  & c
\end{pmatrix} ,  \\
\mapa &(\hx,\rx) = \begin{pmatrix} 
\hx \rx \\
(1-\hx)\rx \\
1-r(x)
\end{pmatrix} .
\end{align*}
{\bf H-loss (hierarchical loss)}: now we assume that the target structure $\G$ is a hierarchical binary tree. Then, $\Edge$ is now the set of directed edges, reflecting a {\it parent} relationship among nodes (each node except the root has one parent). Regarding the labeling, we impose the following property : if an oriented pair $(\nu_i,\nu_j) \in \Edge$, then $y_i \geq y_j$, meaning that a child node cannot be greater that his parent node. The H-loss \cite{cesa2006hierarchical} which measures the length of the common path from the root to the leaves between these assignments is defined as follows:
\begin{equation*}
\Delta_H(\hx,y) = \sum_{i=1}^d c_i 1_{\hx_i \neq y_i} 1_{\hx_{p(i)} = y_{p(i)}} ,
\end{equation*}
where $p(i)$ is the index of the parent of $i$ according to the set of edges $\Edge$, and $c_i$ is a set of positive constants non-increasing on paths from the root to the leaves.

Such a loss can be rewritten under the form: $\Delta_H(\hx,y) = \langle \psi_{wa}(y), C \psi_{wa}(h(x)) \rangle$


\begin{align*}
\mapwa(z) = 
\begin{pmatrix}
z \\
G z 
\end{pmatrix}, \; 
C = \begin{pmatrix}
-2 diag(c) & diag(c) \\
diag(c) & 0 
\end{pmatrix},
\end{align*}

$G$ is the adjacency matrix of the underlying binary tree structure and $c$ the vector of weights defined above. The case of the Hamming loss can also be recovered by choosing: 

\begin{align*}
\mapwa(y) = \begin{pmatrix} y \\ 1-y \end{pmatrix}, \;  \psi_a&(\hx,r(x) ) = \begin{pmatrix} 1-\hx \\ \hx \end{pmatrix}, \\
C &= I_{2d},
\end{align*}
where $I_{2d}$ is the $2d$ identity matrix.

\textbf{Abstention-aware H-loss (Ha-loss):} By mixing the H-loss and the abstention-aware binary classification loss, we get the novel Ha-loss \se{which we define as follows}:
\begin{align}
\label{HAlosseq}
&\Delta_{Ha}(\hx,\rx,y) = \sum_{i=1}^d \underbrace{c_{Ai} 1_{\{ \pred_i = a , \pred_{p(i)} = y_{p(i)}  \} } }_{\text{abstention cost}}  \\ 
& + \underbrace{c_{A_c i} 1_{ \{ \pred_i \neq y_i, \pred_{p(i)} = a \}}  }_{\text{abstention regret}} + \underbrace{c_i 1_{\{ \pred_i \neq y_i , \pred_{p(i)} = y_{p(i)}, a \neq  \pred_i  \} }  }_{\text{misclassification cost}} \nonumber,
\end{align}
where $c_{Ai}$ and $c_{A_c i }$ can be chosen as constants or be function of the predictions.
\secor{This loss is thus adapted}{Thus, we have designed this loss so it is adapted}\secmt{plus revendicatif} to hierarchies where some nodes are known to be hard to predict whereas their children are easy to predict. In this case, the abstention choice can be used at a particular node to pay the \al{\textit{cost}} $c_A$ for predicting its child. If this prediction is still a mistake, the price $c_{A_c i }$ is additionally paid \al{and acts as a \textit{regret} cost penalizing the unnecessary abstention chosen at the parent}. Acting on $c_A$ and $c_{Ac}$ provides a way to control the number of abstentions not only through the risk taken by predicting a given node but also its children. For sake of space, the dot product representation with $\mapwa$ and $\mapa$ of this loss is detailed in the supplementary material.

\subsection{Empirical risk minimization for SOLA}
The goal of SOLA is to learn a pair $(h,r)$ from a i.i.d. (training) sample drawn from a probability distribution $\mathcal{D}$ that minimizes the true risk: 
\begin{align*}
\mathcal{R}(h,r) &= \mathbb{E}_{x,y \sim \mathcal{D}} \; \lossa( \hx,r(x),y ), \\
&= \mathbb{E}_{x,y \sim \mathcal{D}} \; \langle  \mapwa(y), C \mapa(\hx,r(x)) \rangle.
\end{align*}

We notice that this risk can be rewritten as an expected valued over the input variables only:
\begin{align*}\label{pb1}
\mathcal{R}(h,r) &= \mathbb{E}_{x} \; \langle \mathbb{E}_{y | x} \mapwa(y), C\mapa(\hx,r(x)) \rangle.
\end{align*}

This pleads for considering the following surrogate problem:
\begin{itemize}
\item Step 1: we define $g^*(x) = \mathbb{E}_{y | x} \mapwa(y) = \min_{g \in (\X \to \mathbb{R}^q)} \underbrace{\mathbb{E}_{x,y} \| \mapwa(y) - g(x)\|^2}_{\text{surrogate risk}}$. \al{$g^*$ is then the minimizer of a square surrogate risk.}
\item Step 2: we solve the following pre-image or decoding problem:
\begin{align*}
(\predh,\predr) 
&= \argmin_{(y_h,y_r) \in \Y^{H,R}
} \; \langle g^*(x), C\mapa(y_h,y_r ) \rangle,
\end{align*}
\end{itemize}
Solving directly 
the problem above raises some difficulties:

\begin{itemize}
\item In practice, as usual, we do not know the expected value of $\mapwa(y)$ conditioned on $x$: $\mathbb{E}_{y|x} \mapwa(y)$ needs to be estimated from the training sample $\{(x_i,y_i),i=1, \ldots, n\}$. This simple regression problem is referred \se{to} as the learning step and will be solved in the next subsection.
\item The complexity of the $\argmin$ problem will depend on some properties of $\psi_a$. We will refer to this problem as the pre-image and show how to solve it practically at a later stage.
\end{itemize}

These pitfalls, common to all structured output learning problems, can be overcome by substituting a surrogate loss to the target loss and proceeding in two steps: 
\begin{enumerate}
\item Solve the surrogate penalized empirical problem (learning phase): 
\begin{equation}\label{eq:sur}
 \min_{g} \frac{1}{n}\sum_{i=1}^n \| \psi_{wa}(y_i)- g(x_i) \|^2 + \lambda \Omega(g), 
\end{equation}
where $\Omega$ is a penalty function and $\lambda$ a positive parameter. Thus, get a minimizer $\hat{g}$ which is an estimate of $\mathbb{E}_{y|x} \mapwa(y)$.
\item Solve the pre-image or {\it decoding} problem:
\begin{align}
\label{Prob2}
(&\predh ,\predr) = \nonumber\\
&\argmin_{(\hx,\rx) \in \Y^{H,R}
} \; \langle \hat{g}(x), C\mapa(\hx,r(x)) \rangle.
\end{align}
\end{enumerate}

\subsection{Estimation of the conditional density $\mathbb{E}_{y|x} \mapwa(y)$ from training data}

We choose to solve this problem in $\mathcal{H} \subset \mathcal{F}(\X,\mathbb{R}^q)$, a vector-valued Reproducing Kernel Hilbert Space associated to an operator-valued kernel $ K: \X \times \X \to \bmL(\mathbb{R}^q)$. For the sake of simplicity, $K$ is chosen as a decomposable operator-valued kernel with identity: $K(x,x') = Ik(x,x')$ where $k$ is a positive definite kernel on $\X$ and $I$ is the $q \times q$ identity matrix. The penalty is chosen as $\Omega(g)= \| g \|^2_{\bmH}$. This choice leads to the ridge regression problem:
\begin{equation}
\label{learning}
\argmin_{g \in \mathcal{H}} \sum_{i=1}^n \| g(x_i) - \mapwa(y_i) \|^2 + \lambda \| g \|^2_{\bmH},
\end{equation}
that admits a unique and well known closed-form solution \cite{micchelli2005learning,brouard2016input}. 

As $\hat{g}(x)$ is only needed at the prediction stage, within the pre-image to solve, it is important to emphasize the dependency of $\hat{g}(x)$ on the feature vectors $\mapwa(y_i)$:
\begin{equation}\label{ridge}
\hat{g}(x) = \sum_{i=1}^n \alpha_i(x) \mapwa(y_i),
\end{equation}
where $\alpha(x)$ is the following vector: 
\begin{equation}
\alpha(x) =  K_x(\mathbf{K} + \lambda I_{qn})^{-1},
\end{equation}
where $K_x=[K(x,x_1), \ldots, K(x,x_n)]$.
$\mathbf{K}$ is the $qn \times qn$ block matrix such that $\mathbf{K}_{i,j} = K(x_i,x_j)$  and $I_{qn}$ is the identity matrix of the same size. $\alpha_i(x)$ is the block $i$ of $\alpha(x)$.
\section{Learning guarantee for structured losses with abstention}

In this section, we give some statistical guarantees when learning predictors in the framework previously described. To this end, we build on recent results in the framework of Least Squares Loss Surrogate \cite{Ciliberto} that are extended to abstention-aware prediction.
\begin{theorem}
\label{excessbound}
Given the definition of $\lossa$ in \eqref{lossadef},  let us denote $(h,r)$, the pair of predictor and reject functions associated to the estimate $\hat{g}$ obtained by solving the learning problem stated in Eq. \eqref{learning}:
\begin{equation*}
(\hx, \rx) = \argmin_{(y_h,y_r) \in \Y^{H,R}} \langle C \mapa(y_h,y_r),  \hat{g}(x) \rangle.
\end{equation*}
Its true risk with respect to $\lossa$ writes as:
\begin{equation*}
\mathcal{R}(h, r) =  \mathbb{E}_x \langle C \mapa(\hx , \rx),  \mathbb{E}_{y|x} \mapwa(y) \rangle.
\end{equation*}
The optimal predictor $(h^*,r^*)$ is defined as:
\begin{equation*}
(h^*(x), r^*(x)) = \argmin_{(y_h,y_r) \in \Y^{H,R}} \langle C \mapa(y_h,y_r), \mathbb{E}_{y|x} \mapwa(y) \rangle.
\end{equation*}
The excess risk of an abstention aware predictor $(h,r)$: $\mathcal{R}(h,r) - \mathcal{R}(h^\star,r^\star)$ is linked to the estimation error of the conditional density $\mathbb{E}_{y|x} \mapwa(y)$ by the following inequality:
\begin{equation}
\mathcal{R}(h,r) - \mathcal{R}(h^\star,r^\star) \leq 2 c_l \sqrt{\mathcal{L}(\hat{g}) - \mathcal{L}(\mathbb{E}_{y|x} \mapwa(y))},
\end{equation}
where  $\mathcal{L}(g)=\mathbb{E}_{x,y} \| \mapwa(y) - g(x) \|^2$, and $c_l = \|C\| \max_{y_h,y_r \in \Y^{H,R} } \|\psi_a(y_h,y_r)\|_{\mathbb{R}^p}$. 
\end{theorem}
The full proof is given in the Supplements. Close to the one in \cite{Ciliberto}, it is extended by taking the sup of the norm of $\psi_a$ over $\Y^{H,R}$. Moreover when the problem (\ref{learning}) is solved by Kernel Ridge Regression, \cite{Ciliberto} have shown the universal consistency and have obtained 
a generalization bound that still holds in our case since it relies on the result of Theorem \ref{excessbound} only. As a consequence the excess risk of predictors built in the SOLA framework is controlled by the risk suffered at the learning step for which we use \secor{on the shelves}{off the shelf} vector valued regressors with their own convergence guarantees.

  
In the following, we specifically study the pre-image problem in the SOLA framework for a class of output structures that we detail hereafter.

\section{Pre-image for hierarchical structures with Abstention}
\label{specifs}
In what follows we focus on a class of structured outputs that can be viewed as hierarchical objects for which we show how to solve the pre-image problems involved for 
a large class of losses.

\subsection{Hierarchical output structures}
\begin{mydef}
A HEX graph $G = (V,E_h,E_e)$ is a graph consisting of a set of nodes V = $\{v_1,\ldots,v_n\}$, directed edges $E_h \subset V \times V$, and undirected edges $E_e \subset V \times V$ , such that the subgraph $G_h = (V, E_h)$ is a directed acyclic graph (DAG) and the subgraph $G_e = (V, E_e)$ has no self loop.
\end{mydef}

\begin{mydef}
An assignment (state) $y \in \{0, 1\}^d$ of labels $V$ in a HEX graph $G = (V,E_h,E_e)$ is legal if for any pair of nodes labeled $(y_{(i)},y_{(j)}) = (1,1)$, $(v_i,v_j)\notin E_e$ and for any pair $(y_{(i)},y_{(j)}) = (0,1)$, $(v_i,v_j) \notin E_h$. 
\end{mydef}
\begin{mydef}
The state space $SG \subseteq \{0,1\}^d$ of graph $G$ is the set of all legal assignments of $G$.
\end{mydef}

\al{Thus a HEX graph can be described by a pair of (1) a directed graph over a set of binary nodes indicating that any child can be labeled 1 only if its parent is also labeled 1 and (2) an undirected graph of exclusions such that two nodes linked by an edge cannot be simultaneously labeled 1.}
Note that HEX graphs can represent any type of binary labeled graph since $E_h$ and $E_e$ can be empty sets. In previous works, they have been used to model some coarse to fine ontology through the hierarchy $G_h$ while incorporating some prior known labels exclusions encoded by $G_e$ \cite{deng2014large,BenTaieb2016}


While the output data we consider consists of HEX graph assignments 
, our predictions with abstention $(h(x),r(x))$ belong to another space $\Y^{H,R}\subseteq \{0,1\}^d \times \{0,1\}^d$ for which we do not restrict $h(x)$ to belong to $\Y$ but rather allow for other choices detailed in the next section.

\subsection{Efficient solution for the preimage problem}

The complexity of the preimage problem is due to two aspects:
i) the space in which we search the solution ($\Y^{H,R}$) can be hard to explore; and
ii) the $\psi_a$ function can lead to high dimensional representations for which the minimization problem is harder.

The pre-image problem involves a minimization over a constrained set of binary variables. For a large class of abstention-aware predictors we propose a branch-and-bound formulation for which a nearly optimal initialization point can be obtained in a polynomial time. Following the line given by the form of our abstention aware predictor $f^{h,r}$ defined in \secor{section}{Section} \ref{SOLA}, we consider losses involving binary interaction between the predict function $\hx$ and the reject function $\rx$, and suppose that there exists a rectangular matrix $M$ such that $\mapa(\hx,\rx) = M \begin{pmatrix}
\hx \\
\rx \\
\hx \otimes \rx
\end{pmatrix}$ where $\otimes$ is the Kronecker product between vectors. Such a class takes as special cases the examples presented in Section \ref{SOLA}.
We state the following linearization theorem under binary interaction hypothesis:
\begin{theorem}
\label{linearization}
Let $l_{ha}$ be an abstention-aware loss defined by its output mappings $\mapwa$, $\mapa$ and the corresponding cost matrix $C$.

If the $\psi_a$ mapping is a linear function of the binary interactions of $\hx$ and $\rx$ i.e. there exists a matrix $M$ such that $ \forall (\hx,\rx) \in \Y^{H,R} \; \mapa(\hx,\rx) = M \begin{pmatrix}
\hx \\
\rx \\
\hx \otimes \rx
\end{pmatrix} $, then there exists a bounded linear operator $A$ and a vector $b$ such that $\forall \psi_x \in \mathbb{R}^p$ the pre-image problem:
\begin{align*}
(\predh,\predr) =  \argmin_{(y_h,y_r) \in \Y^{H,R} }  \langle   \mapa(y_h,y_r),\psi_x \rangle,
\end{align*}
has the same solutions as the linear program:
\begin{align*}
\predh,\predr = &  \argmin_{(y_h,y_r) \in \Y^{H,R}} [y_h^T y_r^T c^T] M^T \psi_x \\
& \text{s.t.} \; A \begin{pmatrix}
y_h \\
y_r \\
c
\end{pmatrix} \leq b.
\end{align*}
Where $c$ is a $d^2$ dimensional vector constrained to be equal to $y_h \otimes y_r$.
\end{theorem}
The proof is detailed in the supplementary material.


The problem above still involves a minimization over the structured binary set $\Y^{H,R}$. Such a set of solutions encodes some predefined constraints:

\begin{itemize}
\item Since the objects we intend to predict are HEX graph \al{assignments}, the vectors of the output space $y \in \Y$ should satisfy the hierachical constraint : $y_i \leq y_{p(i)}$ with $p(i)$ the index of the parent of $i$ according to the hierarchy. When predicting with abstention we relax this condition since we suppose that a descendant node can take the value $y_i = 1$ if its parent was active $y_{p(i)} = 1$ or if we abstained from predicting it $r_{p(i)} = 0$. Such a condition is equivalent to the constraint 
\begin{equation}
y_i  r_{p(i)} \leq y_{p(i)} r_{p(i)}.
\end{equation}

\item A second condition we used in practice is the restriction of the use of abstention for two consecutive nodes: structured abstention at a layer must be used in order to reveal a subsequent prediction which is known to be easy. Such a condition can be encoded through the inequality:
\begin{equation}
r_i + r_{p(i)} \leq 1.
\end{equation}
\end{itemize}

In our experiments, the structured space $\Y^{H,R}$ has been chosen as the set of binary vectors $(\hx,\rx) \in \Y^{H,R}$ that respect the two above conditions. These choices are motivated by our application but note that any subset of $\{0,1\}^d \times \{0,1\}^d$ can be built in a similar way by adding some inequality constraints: 
$A_{\Y^{H,R}} \begin{pmatrix} \hx \\ \rx \\ \hx \otimes \rx \end{pmatrix} \leq b_{\Y^{H,R}}$. Consequently, the $\Y^{H,R}$ constraints can be added to the previous minimization problem to build the canonical form:
\begin{align*}
(\predh,\predr) = &  \argmin_{(y_h,y_r)}  [y_h^T y_r^T c^T] M^T \psi_x \\
& \text{s.t.} \; A_\text{canonical} \begin{pmatrix}
y_h \\
y_r \\
c
\end{pmatrix} \leq b_\text{canonical}, \\
& (y_h,y_r) \in \{0,1\}^d \times \{0,1\}^d,
\end{align*}
where $A_\text{canonical} = \begin{pmatrix} A \\ A_{\Y^{H,R}} \end{pmatrix}$ and $b_\text{canonical} = \begin{pmatrix} b \\ b_{\Y^{H,R}} \end{pmatrix}$.

The complexity of the problem above is linked to some properties of the $A_\text{canonical}$ operator. \cite{goh} have shown that in the case of the minimization of the H-loss with hierarchical constraints, the linear operator $A_\text{canonical}$ satisfies the property of total unimodularity \cite{schrijver1998theory} which is a sufficient condition for the problem above to have the same solutions as its continuous relaxation leading to a polynomial time algorithm. In the \al{more} general case \al{of the Ha-loss}, solving such an integer program is NP-hard and the optimal solution can be obtained using a branch-and-bound algorithm. When implementing this type of approach, the choice of the initialization point can strongly influence the convergence time. \al{As in practical applications, we expect the number of abstentions to remain low}, such a point can be chosen as the solution of the original prediction problem without abstention \cite{goh}. Moreover since the abstention mechanism should modify only a small subset of the predictions, we expect this solution to be close to the abstention aware one. 
\section{Numerical Experiments}
We study three subtasks of opinion mining, namely sentence-based aspect prediction, sentence-based joint prediction of aspects and polarities (possibly with abstention) and full review-based star rating. We show that these tasks can be linked using a hierarchical graph similar to the probabilistic model of \cite{marcheggiani2014hierarchical} and exploit the abstention mechanism to build a robust pipeline: based on the opinion labels available at the sentence-level, we build a two-stage predictor that first predicts the aspects and polarities at the sentence level, before deducing the corresponding review-level values. 
\subsection{Parameterization of the Ha-loss}
In all our experiments, we rely on the expression of the Ha-loss presented in \ref{HAlosseq}.
The linear programming formulation \al{of the pre-image problem} used in the branch-and-bound solver is derived in the supplementary material and involves a decomposition similar to the one described in Section \ref{SOLA} for the H-loss. 
Implementing the Ha-loss requires choosing the weights $c_i, c_{Ai}$ and $c_{A_c i}$. We first fix the $c_i$ weights in the following way :
\begin{align*}
c_0 &= 1 \\
c_i &= \frac{c_{p(i)}}{|\text{siblings(i)|}} \; \forall i \in \{1, \ldots, d\}.
\end{align*}
Here, $0$ is assumed to be the index of the root node. This weighting scheme has been commonly used in previous studies \cite{rousu2006kernel,bi2012hierarchical} and is related to the minimization of the Hamming Loss on a vectorized representation of the graph assignment. As far as the abstention weights $c_{Ai}$ and $c_{A_c i}$ are concerned, making an exhaustive analysis of all the possible choices is impossible due to the number of parameters involved. Therefore, our experiments focus on weighting schemes built in the following way:
\begin{align*}
c_{Ai} &= K_A c_i \\
c_{A_c i} &= K_{A_c} c_i
\end{align*}
The effect of the choices of $K_A$ and $K_{A_c}$ will be illustrated below on the opinion prediction task. We also ran a set of experiments on a hierarchical classification task of MRI images from the IMAGECLEF2007 dataset reusing the setting of \cite{Dimitrovski08:proc} where we show the results obtained for different $c_i$ weighting schemes. The settings and the results have been placed in the supplementary material.
\subsection{Learning with Abstention for aspect-based opinion mining}
\begin{figure}[h]
\includegraphics[width = \linewidth]{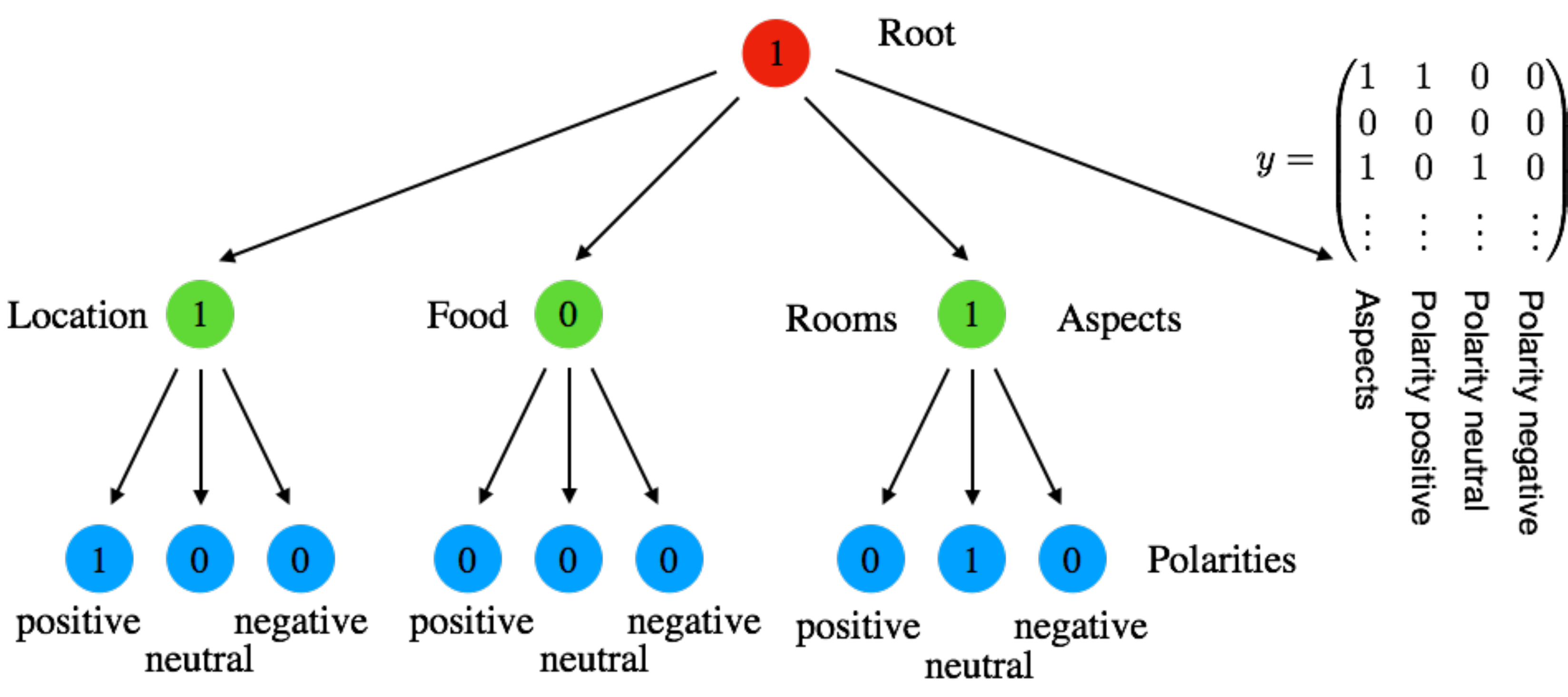}
\caption{Graphical representation of the opinion structure}
\label{graph_opinion}
\end{figure}

We test our model on the problem of aspect-based 
opinion mining on a subset of the TripAdvisor dataset released in \cite{marcheggiani2014hierarchical}. It consists of 
369 hotel reviews for a total of 4856 sentences with predefined train and test sets. In addition to the review-level star ratings, the authors gathered the opinion annotations at the sentence-level for a set of 11 predefined aspects and their corresponding polarity. Similarly to them, we discard the ``NOT RELATED'' aspect and consider the remaining 10 aspects with the 3 different polarities (positive, negative or neutral) for each. We propose a graphical representation of the opinion structure at the sentence level (see Fig.~\ref{graph_opinion}). Objects in the output space $y\in \mathcal{Y}$ consist of 
trees of depth 3 where the first node is the root, the second layer is made of aspect labels and the third one is the polarities corresponding to each aspect. The corresponding assignments are encoded by a binary matrix $y \in \Y$ where $y$ is the concatenation of the vectors indicating the presence of each aspect (depth 2) and the ones indicating the polarity. 

An example of $y$ encoding is displayed in Fig.\ref{graph_opinion}.  
Based on the recent results of \cite{DBLP:journals/corr/ConneauKSBB17}, we focus on the InferSent representation to encode our inputs. This dense sentence embedding corresponds to the inner representation of a deep neural network trained on a natural language inference task and has been shown to give competitive results in other natural language processing tasks.

We test our model on 3 different subtasks. In \textbf{Exp1}, we first apply our model (H Regression InferSent) to the task of opinion aspect prediction  and compare it against two baselines and the original results of \cite{marcheggiani2014hierarchical}.
In \textbf{Exp2}, we test our method and baselines on the problem of joint aspect and polarity prediction in order to assess the ability of the hierarchical predictor to take advantage of the output structure. On this task we additionally illustrate the behavior of abstention when varying the constants $K_A$ and $K_{A_c}$.
In \textbf{Exp3}, we illustrate the use abstention as a mean to build a robust pipeline on the task of star rating regression based on a sentence-level opinion predictor.

\textbf{Exp1. Aspect prediction.\ }
In this first task, we aim at predicting the different aspects discussed in each sentence. This problem can be cast as a multilabel classification problem where the target is the first column of the output objects $y$ for which we devise two baselines. The first relies on a logistic regression  model (Logistic Regression InferSent) trained \se{separately} for each aspect. The second baseline (Linear chain Conditional Random Fields (CRF) \cite{sutton2012introduction} InferSent) is inspired by the work of \cite{marcheggiani2014hierarchical} who built a hierarchical CRF model based on a handcrafted sparse feature set including one-hot word encoding, POS tags and sentiment vocabulary. Since the optimization via Gibbs sampling of their model relies on the sparsity of the feature set, we could not directly use it with our dense representation. 
Linear chain CRF InferSent takes advantage of our input features while remaining computationally tractable. One linear chain is trained for each node of the output structures and the chain encodes the dependency between successive sentences.

Table \ref{Table1} below shows the results in terms of micro-averaged F1 ($\mu \text{-F1}$) score obtained on the task of aspect prediction.
\begin{table}[h]
\centering
\label{Table1}
\begin{tabular}{lrr}
\hline
method                                   &  $\mu \text{-F1}$    \\ \hline
H Regression InferSent                  &     0.59                        \\ \hline
Logistic Regression InferSent    & 0.60 \\ \hline
Linear chain CRF InferSent        &       0.59                   \\ \hline
\begin{tabular}[c]{@{}l@{}}
Linear chain CRF sparse features   \\ \citeauthor{marcheggiani2014hierarchical}
\end{tabular}                                    &   0.49                        \\ \hline
\begin{tabular}[c]{@{}l@{}}Hierarchical CRF sparse features \\ \citeauthor{marcheggiani2014hierarchical}
\end{tabular}                                   &   0.49                        \\ \hline
\end{tabular}
\caption{Experimental results on the TripAdvisor dataset for the aspect prediction task.}
\end{table}
The three methods using InferSent give significantly better results than \cite{marcheggiani2014hierarchical}. Consequently,  the next experiments will not consider them. Even though  H Regression was trained in order to predict the whole structure, it obtains results similar to logistic regression and linear chain CRF.

\textbf{Exp2. Joint polarity and aspect prediction with abstention.\ }We take as output objects the assignments of the graph described (Fig. \ref{graph_opinion}) and build an adapted abstention mechanism. Our intuition is that in some cases, the polarity might be easier to predict than the aspect to which it is linked. This can typically happen when some vocabulary linked to the current aspect has been unseen during the training or is implicit whereas the polarity vocabulary is correctly recognized. An example is the sentence " We had great views over the East River" where the aspect "Location" is implicit and where the "views" could mislead the predictor and result in a prediction of the aspect "Other". In such a case, \cite{marcheggiani2014hierarchical} underline that the inter-annotator agreement is low. 
For this reason, we want that our classifier allows multiple candidates for aspect prediction while providing the polarity corresponding to them. 
We illustrate this behavior by running two sets of experiments in which we do not allow the predictor to abstain on the polarity.

\begin{figure}[h]
\includegraphics[width = \linewidth]{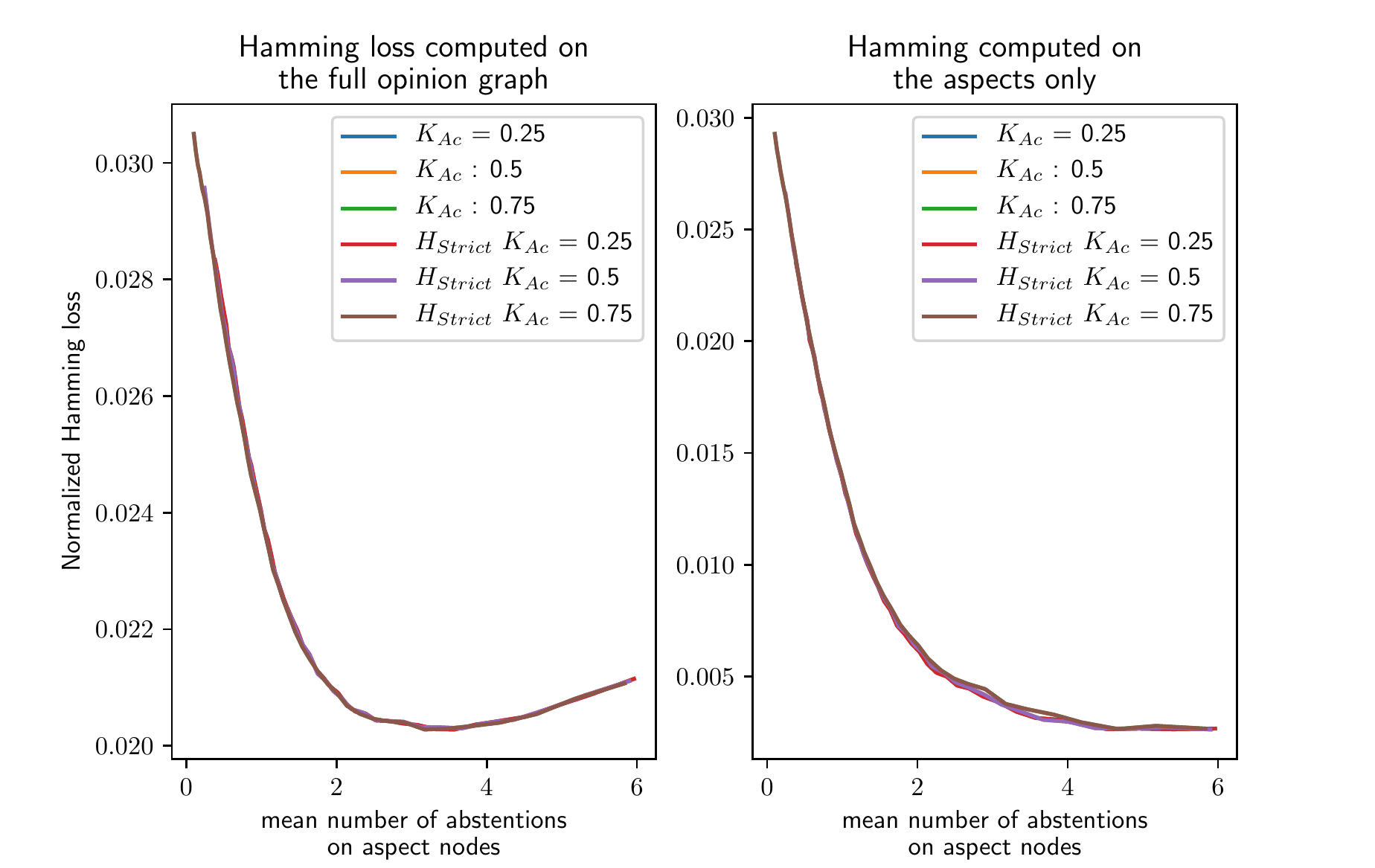}
\caption{Hamming loss as a function of the number of aspect labels where the predictor abstained itself.}
\label{Abstention_aspects}
\end{figure}

In the first experiment, we want to analyze the influence of the parameterization of the Ha-loss. Following the parameterization of $c_{Ai}$ and $c_{A_c i}$ previously proposed, we generated some predictions with varying values of $K_A \in [0,0.5]$ and $K_{A_c} \in \{0.25,0.5,0.75\}$. We displayed the Hamming loss between the true labels and the predictions as a function of the mean number of aspects on which the predictor abstained (Fig.~\ref{Abstention_aspects}) and handle two cases : \al{modified :} in the left figure, all nodes except the one on which we abstained were used to compute the Hamming loss. In the right one, all nodes except the aspect on which we abstained and their corresponding polarity were used to compute the Hamming loss.
The $H_\text{Strict}$ results correspond to a predictor for which the original hierarchical constraint is forced: $y_{(i)}\leq y_{p(i)}$ and the three other curves 
have been obtained with the generalized constraint hypothesis $y_{(i)} r_{p(i)} \leq y_{p(i)} r_{p(i)}$. 

We additionally ran our model H Regression without abstention and our two baselines logistic regression for which we measured a similar Hamming loss of 0.03 (corresponding to 0 abstention on the left Figure \ref{Abstention_aspects}). Concerning the micro-averaged F1 score, the H Regression retrieved a score of $0.54$ being slightly above the logistic regression which scored $0.53$ and the linear chain CRF with $0.52$. 

Two conclusions can be raised. Firstly, the value of $K_{A_c}$ and the choice of the hypothesis $H_\text{Strict}$ have little to no influence on the scores computed in the two cases previously described. Secondly, increasing the number of abstentions on aspects 
helps reducing the number of errors counted on the aspects nodes when the predictor abstains on less than 3 labels. After this point, the quality of the overall prediction decreases since the error rate on the remaining aspects selected for abstention is less than the one on the polarity labels

\begin{figure}[h]
\includegraphics[width = \linewidth]{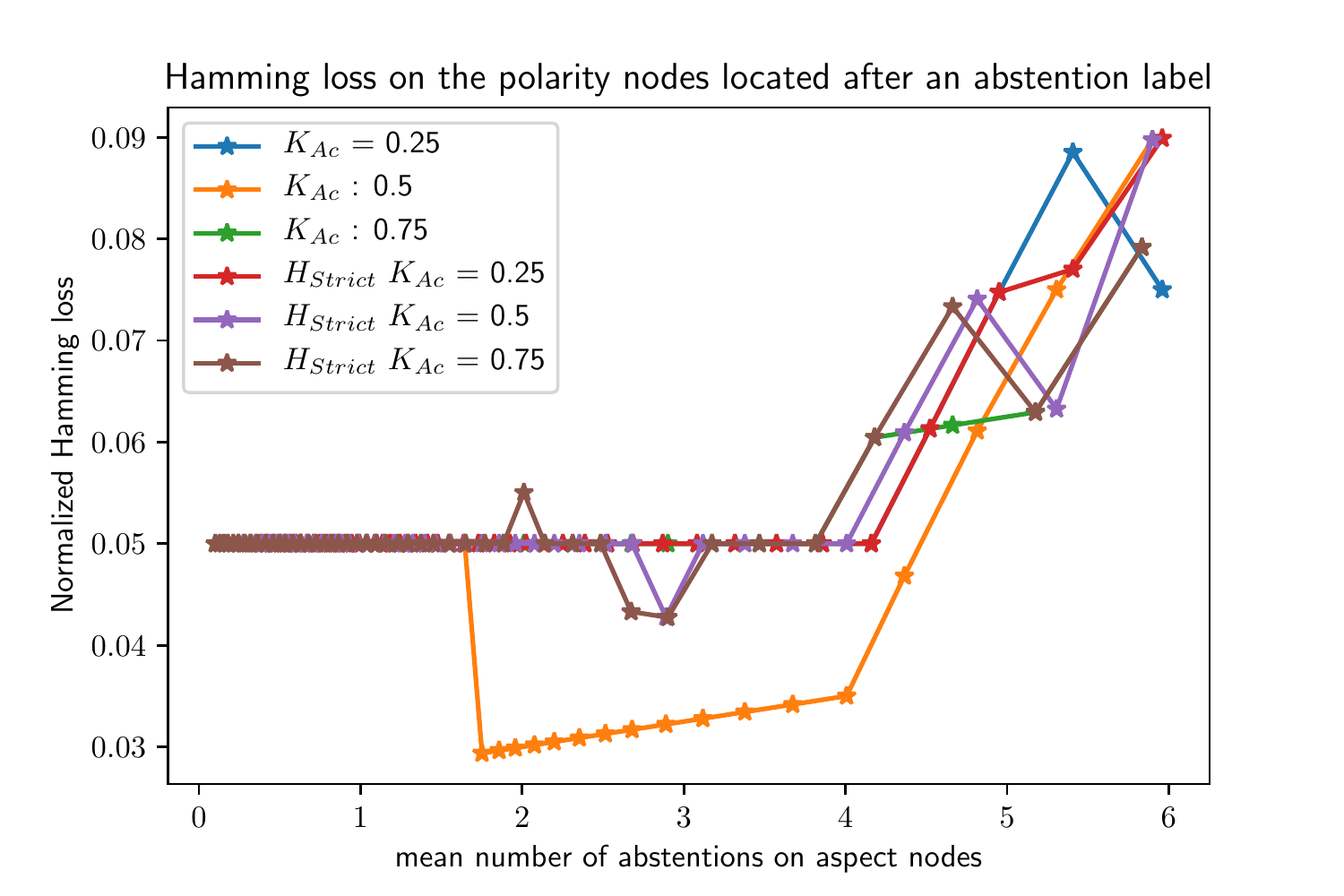}
\caption{Hamming loss computed on polarity nodes located after an aspect for which the predictor abstained}
\label{Hammingpolarity}
\end{figure}

Subsequently, we examine the Hamming loss on the polarity predictions situated after an aspect node to understand the influence of the $c_{A_c}$ coefficients and the relaxation of the $H_\text{Strict}$ hypothesis in Figure \ref{Hammingpolarity}.
The orange curve gives the best score when the mean number of abstentions is between 2 and 4 per sentence. The only difference with the $H_\text{strict}$ hypothesis is the ability to predict the polarity of an aspect candidate for abstention even if the predictor function does not select it. This behavior is made possible by the fact that our prediction does not respect the $\Y$ constraints but instead belong to the more flexible space $\Y^{H,R}$ 
Finally we show how abstention can be used to build a robust pipeline for star-rating regression.

\textbf{Exp3. Star rating regression at the review level based on sentence level predictions.\ } In the last round of experiments, we show that abstention can be used as a way to build a robust intermediate representation for the task of opinion rating regression \cite{wang2011latent} which consists in predicting the overall average star rating given by each reviewer on a subset of six predefined aspects. The figure below illustrates the different elements involved in our problem.
\begin{figure}[h]
\includegraphics[width = \linewidth]{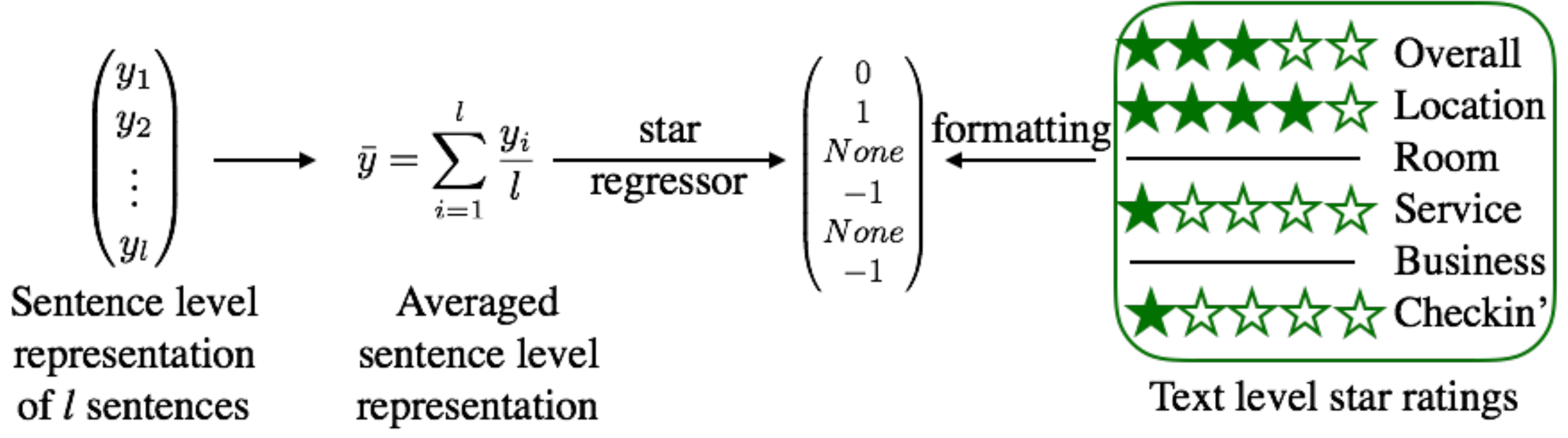}
\caption{Star rating regression pipeline}
\label{pipeline}
\end{figure}
The procedure is split in two steps. Firstly, we learn a sentence-level opinion predictor that takes advantage of the available annotations. This step corresponds to the one studied in the previous experiment. Then a vector-valued regressor (star regressor in Figure~\ref{pipeline}) is built. 
It takes as input the component-wise average of the sentence level opinion representations
, and intends to predict the star ratings at the review level. For each of the five overall aspects a separate Ridge Regressor is trained based on the true labels available. Once learned, the regressors take as input the prediction of the first step in a pipelined way 

Similarly to \cite{marcheggiani2014hierarchical}, we rescale the star ratings on a (-1,0,1) 
scale and report the macro-averaged mean average error on the test-set in Table \ref{Table2} below under the column MAE text level. We additionally include the MAE error measured on polarity predictions at the sentence level counted when the underlying aspect predicted is a true positive.
\begin{table}[h]
\centering
\label{Table2}
\begin{tabular}{lcr}
\hline
method                                  & \begin{tabular}[c]{@{}c@{}}$MAE$ \\ sentence level\end{tabular} & \begin{tabular}[c]{@{}c@{}}$MAE$ \\ text level \end{tabular}   \\ \hline
\begin{tabular}[c]{@{}c@{}}
Oracle: regression with \\ true sentence labels \end{tabular} & 0 & 0.38   \\ \hline \hline
\begin{tabular}[c]{@{}c@{}}Hierarchical CRF \\ 
\end{tabular}                         & 0.50     &        0.50     \\ \hline
H Regression                &     0.30 & 0.45                 \\ \hline
\begin{tabular}[c]{@{}c@{}}H Regression \\ with Abstention\end{tabular} &  C                           & 0.43 \\ \hline
\end{tabular}
\caption{Experimental result on the TripAdvisor dataset for the polarity prediction task}
\end{table}
The first row is our oracle: the sentence-level opinion representations are assumed to be known on the test set and fed in the text-level opinion regressors to find back the star ratings. The Hierarchical CRF line corresponds to the best results reported by \cite{marcheggiani2014hierarchical} on the two tasks. H Regression is our model without abstention used as a predictor of the sentence-level representation in the pipeline shown in Fig~\ref{pipeline}. Finally for the H Regression with abstention, we used as a sentence-level representation : $y_\text{a} = \hx - (1-\rx)$. Since the only non-zero components of $(1-\rx)$ correspond to aspects on which we abstained, subtracting them from the original prediction results in a reduction of the confidence of the regressor for these aspects and biasing the corresponding polarity predictions towards 0. H Regression strongly outperforms Hierarchical CRF on both tasks. We do not report the score for H Regression with abstention since it is dependent on the number of abstentions but show that it improves the results of the H Regression model on the text-level prediction task. \al{The significance of the scores has been assessed with a Wilcoxon rank sum test (p-value $10^-6$).}

\section{Conclusion}

The novel framework, Structured Learning with Abstention, extends two families of approaches: learning with abstention and least-squares surrogate structured prediction. 
It is important to notice that beyond ridge regression, any vector-valued regression model that writes as \eqref{ridge} is eligible. This is typically the case of Output Kernel tree-based methods \cite{GeurtsWd06}. Also, SOLA has here been applied to opinion analysis but it could prove suitable for more complex structure-labeling problems.
Concerning Opinion Analysis, we have shown that abstention can be used to build a robust representation for star rating in a pipeline framework. One extension of our work would consist in 
learning how to abstain by jointly predicting the aspects and polarity at the sentence and text level.
\section*{Acknowledgements}

This work has been funded by the french ministry of research and by the chair Machine Learning for Big Data of Télécom ParisTech.

\bibliography{biblio}

\begin{thebibliography}{20}
\providecommand{\natexlab}[1]{#1}
\providecommand{\url}[1]{\texttt{#1}}
\expandafter\ifx\csname urlstyle\endcsname\relax
  \providecommand{\doi}[1]{doi: #1}\else
  \providecommand{\doi}{doi: \begingroup \urlstyle{rm}\Url}\fi

\bibitem[BenTaieb \& Hamarneh(2016)BenTaieb and Hamarneh]{BenTaieb2016}
BenTaieb, A. and Hamarneh, G.
\newblock \emph{Topology Aware Fully Convolutional Networks for Histology Gland
  Segmentation}, pp.\  460--468.
\newblock Springer International Publishing, Cham, 2016.
\newblock ISBN 978-3-319-46723-8.
\newblock \doi{10.1007/978-3-319-46723-8_53}.
\newblock URL \url{https://doi.org/10.1007/978-3-319-46723-8_53}.

\bibitem[Bi \& Kwok(2012)Bi and Kwok]{bi2012hierarchical}
Bi, W. and Kwok, J.~T.
\newblock Hierarchical multilabel classification with minimum bayes risk.
\newblock In \emph{Data Mining (ICDM), 2012 IEEE 12th International Conference
  on}, pp.\  101--110. IEEE, 2012.

\bibitem[Brouard et~al.(2016)Brouard, Szafranski, and d’Alch{\'e}
  Buc]{brouard2016input}
Brouard, C., Szafranski, M., and d’Alch{\'e} Buc, F.
\newblock Input output kernel regression: supervised and semi-supervised
  structured output prediction with operator-valued kernels.
\newblock \emph{Journal of Machine Learning Research}, 17\penalty0
  (176):\penalty0 1--48, 2016.

\bibitem[Cesa-Bianchi et~al.(2006)Cesa-Bianchi, Gentile, and
  Zaniboni]{cesa2006hierarchical}
Cesa-Bianchi, N., Gentile, C., and Zaniboni, L.
\newblock Hierarchical classification: combining bayes with svm.
\newblock In \emph{Proceedings of the 23rd international conference on Machine
  learning}, pp.\  177--184. ACM, 2006.

\bibitem[Ciliberto et~al.(2016)Ciliberto, Rosasco, and Rudi]{Ciliberto}
Ciliberto, C., Rosasco, L., and Rudi, A.
\newblock A consistent regularization approach for structured prediction.
\newblock In Lee, D.~D., Sugiyama, M., Luxburg, U.~V., Guyon, I., and Garnett,
  R. (eds.), \emph{Advances in Neural Information Processing Systems 29}, pp.\
  4412--4420. Curran Associates, Inc., 2016.

\bibitem[Clavel \& Callejas(2016)Clavel and Callejas]{clavel2016sentiment}
Clavel, C. and Callejas, Z.
\newblock Sentiment analysis: from opinion mining to human-agent interaction.
\newblock \emph{IEEE Transactions on affective computing}, 7\penalty0
  (1):\penalty0 74--93, 2016.

\bibitem[Conneau et~al.(2017)Conneau, Kiela, Schwenk, Barrault, and
  Bordes]{DBLP:journals/corr/ConneauKSBB17}
Conneau, A., Kiela, D., Schwenk, H., Barrault, L., and Bordes, A.
\newblock Supervised learning of universal sentence representations from
  natural language inference data.
\newblock \emph{CoRR}, abs/1705.02364, 2017.
\newblock URL \url{http://arxiv.org/abs/1705.02364}.

\bibitem[Cortes et~al.(2016)Cortes, DeSalvo, and Mohri]{CortesAbstention}
Cortes, C., DeSalvo, G., and Mohri, M.
\newblock Boosting with abstention.
\newblock In Lee, D.~D., Sugiyama, M., Luxburg, U.~V., Guyon, I., and Garnett,
  R. (eds.), \emph{Advances in Neural Information Processing Systems 29}, pp.\
  1660--1668. Curran Associates, Inc., 2016.

\bibitem[Deng et~al.(2014)Deng, Ding, Jia, Frome, Murphy, Bengio, Li, Neven,
  and Adam]{deng2014large}
Deng, J., Ding, N., Jia, Y., Frome, A., Murphy, K., Bengio, S., Li, Y., Neven,
  H., and Adam, H.
\newblock Large-scale object classification using label relation graphs.
\newblock In \emph{European Conference on Computer Vision}, pp.\  48--64.
  Springer, 2014.

\bibitem[Dimitrovski et~al.(2008)Dimitrovski, Kocev, Loskovska, and
  D\v{z}eroski]{Dimitrovski08:proc}
Dimitrovski, I., Kocev, D., Loskovska, S., and D\v{z}eroski, S.
\newblock Hierchical annotation of medical images.
\newblock In \emph{Proceedings of the 11th International Multiconference -
  Information Society IS 2008}, pp.\  174--181. IJS, Ljubljana, 2008.

\bibitem[Geurts et~al.(2006)Geurts, Wehenkel, and
  d'Alch{\'{e}}{-}Buc]{GeurtsWd06}
Geurts, P., Wehenkel, L., and d'Alch{\'{e}}{-}Buc, F.
\newblock Kernelizing the output of tree-based methods.
\newblock In \emph{Machine Learning, Proceedings of the Twenty-Third
  International Conference {(ICML} 2006), Pittsburgh, Pennsylvania, USA, June
  25-29, 2006}, pp.\  345--352, 2006.

\bibitem[{Goh} \& {Jaillet}(2016){Goh} and {Jaillet}]{goh}
{Goh}, C.~Y. and {Jaillet}, P.
\newblock {Structured Prediction by Conditional Risk Minimization}.
\newblock \emph{ArXiv e-prints}, November 2016.

\bibitem[Marcheggiani et~al.(2014)Marcheggiani, T{\"a}ckstr{\"o}m, Esuli, and
  Sebastiani]{marcheggiani2014hierarchical}
Marcheggiani, D., T{\"a}ckstr{\"o}m, O., Esuli, A., and Sebastiani, F.
\newblock Hierarchical multi-label conditional random fields for
  aspect-oriented opinion mining.
\newblock In \emph{ECIR}, pp.\  273--285. Springer, 2014.

\bibitem[Micchelli \& Pontil(2005)Micchelli and Pontil]{micchelli2005learning}
Micchelli, C.~A. and Pontil, M.
\newblock On learning vector-valued functions.
\newblock \emph{Neural computation}, 17\penalty0 (1):\penalty0 177--204, 2005.

\bibitem[Osokin et~al.(2017)Osokin, Bach, and Lacoste{-}Julien]{osinski2017}
Osokin, A., Bach, F.~R., and Lacoste{-}Julien, S.
\newblock On structured prediction theory with calibrated convex surrogate
  losses.
\newblock In \emph{Advances in Neural Information Processing Systems 30}, pp.\
  301--312, 2017.

\bibitem[Pontiki et~al.(2016)Pontiki, Galanis, Papageorgiou, Androutsopoulos,
  Manandhar, Mohammad, Al-Ayyoub, Zhao, Qin, De~Clercq,
  et~al.]{pontiki2016semeval}
Pontiki, M., Galanis, D., Papageorgiou, H., Androutsopoulos, I., Manandhar, S.,
  Mohammad, A.-S., Al-Ayyoub, M., Zhao, Y., Qin, B., De~Clercq, O., et~al.
\newblock Semeval-2016 task 5: Aspect based sentiment analysis.
\newblock In \emph{Proceedings of the 10th international workshop on semantic
  evaluation (SemEval-2016)}, pp.\  19--30, 2016.

\bibitem[Rousu et~al.(2006)Rousu, Saunders, Szedmak, and
  Shawe-Taylor]{rousu2006kernel}
Rousu, J., Saunders, C., Szedmak, S., and Shawe-Taylor, J.
\newblock Kernel-based learning of hierarchical multilabel classification
  models.
\newblock \emph{Journal of Machine Learning Research}, 7\penalty0
  (Jul):\penalty0 1601--1626, 2006.

\bibitem[Schrijver(1998)]{schrijver1998theory}
Schrijver, A.
\newblock \emph{Theory of linear and integer programming}.
\newblock John Wiley \& Sons, 1998.

\bibitem[Sutton et~al.(2012)Sutton, McCallum, et~al.]{sutton2012introduction}
Sutton, C., McCallum, A., et~al.
\newblock An introduction to conditional random fields.
\newblock \emph{Foundations and Trends{\textregistered} in Machine Learning},
  4\penalty0 (4):\penalty0 267--373, 2012.

\bibitem[Wang et~al.(2011)Wang, Lu, and Zhai]{wang2011latent}
Wang, H., Lu, Y., and Zhai, C.
\newblock Latent aspect rating analysis without aspect keyword supervision.
\newblock In \emph{Proceedings of the 17th ACM SIGKDD international conference
  on Knowledge discovery and data mining}, pp.\  618--626. ACM, 2011.

\end{thebibliography}
\bibliographystyle{icml2018}

\end{document}